\documentclass[cameraready]{Interspeech}

\title{Robustness Assessment of Large Audio Language Models\\ in Multiple-choice Evaluation}

\author[affiliation={1, 2}, orcid=0000-0002-7705-2250, correspondingauthor]{Fernando}{López}
\author[affiliation={3}, orcid=0000-0002-3725-742X,]{Santosh}{Kesiraju}
\author[affiliation={1}, orcid=0000-0002-4507-4930,]{Jordi}{Luque}

\address{
    $^1$ Scientific Research, Telefónica Innovación Digital, Spain \\
    $^2$ Universidad Autónoma de Madrid, Spain \\
    $^3$ Brno University of Technology, Czech Republic
}

\email{fernando.lopez@telefonica.com}


\usepackage{comment, multirow}

\begin{document}

\maketitle

\begin{abstract}
    Recent advances in large audio language models (LALMs) are primarily assessed using multiple-choice question answering (MCQA). However, subtle changes such as shifting choice order yield substantially different results. Moreover, textual questions and options contain linguistic hints that let models infer correct answers without relying on the audio. Existing MCQA frameworks ignore these concerns and report a single accuracy number. We dive into MCQA evaluation and conduct a systematic study spanning three benchmarks (MMAU, MMAR, MMSU) and four models: Audio Flamingo 2, Audio Flamingo 3, Qwen2.5-Omni-7B-Instruct, and Kimi-Audio-7B-Instruct. Our findings show that language bias is present across benchmarks, and models are sensitive not only to choice ordering but also to paraphrasing of questions and options. Finally, we propose a simpler evaluation protocol and metric that account for these variations and provide a more detailed assessment of LALMs within the MCQA framework. 
\end{abstract}

\section{Introduction}
\label{sec:intro}
Understanding auditory information is essential for fostering natural human-machine interactions. Following the rapid advancement of multimodal large language models (MLLMs), the field of large audio language models (LALMs) has emerged quickly. Recently, numerous LALMs have appeared, including LTU \cite{gong2023listen}, SALMONN \cite{sun2024video}, GAMA \cite{ghosh2024gama}, Audio Flamingo 2 \cite{ghoshaudio}, Qwen2.5-Omni \cite{xu2025qwen2}, Audio Reasoner \cite{xie2025audio}, Kimi-Audio \cite{ding2025kimi}, and Audio Flamingo 3 \cite{goel2025audioflamingo3advancing}. Evaluation of these models is critical for ranking their performance and identifying limitations, thereby facilitating advancement in the field. To this end, several benchmarks have appeared, initially focusing primarily on foundational tasks (e.g., speech recognition) \cite{huang2024dynamic, yang2024air, wang2024audiobench, huang2024dynamic2}. More recently, additional benchmarks have begun to incorporate reasoning abilities, such as MMAU \cite{sakshi2024mmau}, MMAR \cite{ma2025mmar}, SAKURA \cite{yang2025sakura}, MMSU \cite{wang2025mmsu}, and MMAU-Pro \cite{kumar2026mmau}.

A significant portion of these benchmarks operate within the multiple-choice question-answering (MCQA) framework. This approach is well-established in the LLM domain \cite{zheng:2024:LLM_MCQ, hendrycks2020measuring, zellers2019hellaswag, wang2024mmlu} due to its simplicity and alignment with human testing practices \cite{zheng:2024:LLM_MCQ}. As a consequence, a considerable amount of LALM evaluations use MCQA, often resulting in a single performance metric for each benchmark or category. However, as studied for the LLM realm, this approach is sensitive to minor modifications, such as changes in the order of options or variations in prompts \cite{zheng:2024:LLM_MCQ, balepur-etal-2025-best, nalbandyan-etal-2025-score}. Moreover, some MCQA items may be solvable from the question and answer options text (language bias). Benchmarks are the primary instrument through which the field tracks progress; if accuracy does not capture genuine model capability, reported improvements may be an artifact of evaluation fragility rather than true advancement. Thus, to understand the true abilities and limitations of the models, it is crucial to investigate these effects in the context of LALMs.

We address this gap by asking: when keeping the audio fixed, do LALMs produce consistent answers under linguistic variations? While robustness to signal-level perturbations (e.g., silence insertion, acoustic event reordering, or duration variation  \cite{bhattacharya2025benchmarking}) 
is important, it is an orthogonal direction. We focus on linguistic sensitivity in MCQA-based multimodal evaluation, analyzing whether audio provides a stabilizing grounding signal or whether linguistic framing dominates.


In this study, we present a comprehensive evaluation framework tailored to capture subtle variations that can impact MCQA outcomes for LALMs. Our methodology begins by establishing baseline performance on existing benchmarks under default conditions, alongside measuring the performance of text-only LLMs as a proxy for language bias. We then implement controlled variations and assess the impact in performance of the models. We evaluate accuracy, consistency, and the model's ability to consistently deliver correct answers across varying conditions. This assessment is accomplished through the systematic application of isolated variations across several factors, including choice ordering, question phrasing, answer phrasing, and distractor phrasing. In addition, we investigate the effects of combining all variations to better understand their impact. Our objective is to establish a more reliable assessment framework for LALMs. 


The contributions of this work are: (i) language bias quantification: some text-only LLMs exceed random chance on all evaluated benchmarks, achieving 48.3\% accuracy (22.6 points above chance). (ii) Demonstrating that choice ordering or the wording of questions, ground-truth answers, or distractors can result in huge performance shifts. Our analysis specifically reveals that altering the phrasing of distractor choices can produce an accuracy standard deviation of up to 13.7\%. (iii) Investigating biases associated with the preference for longer answer candidates. (iv) Introducing a mixed-perturbation setting that captures models' sensitivity at a practical computational cost to account for these critical factors. We evaluate using the correctness rate (CoR), which effectively captures the model variability in performance. This framework offers a reliable method to effectively assess the robustness of LALMs. We release evaluation code and perturbed benchmarks\footnote{\url{https://github.com/ferugit/mcqa-lalms-robustness}.}.


\begin{figure*}[th!]
    \centering
    \includegraphics[width=\textwidth, clip, trim=0 1 0 0, clip, ]{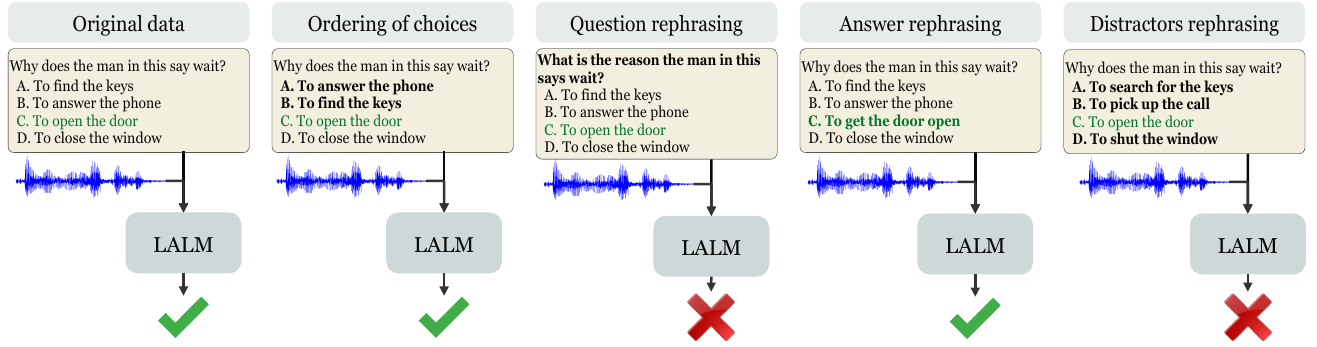}
    \vspace{-0.7cm}
    \caption{Evaluation protocol. Sample of each perturbation applied to a benchmark in isolation. Correct answer is in green.}
    \label{fig:metodology}
\end{figure*}

\section{Methodology}
\label{sec:methodology}
In our methodology, we modify an existing benchmark by altering several key elements: the ordering of choices, phrasing of questions, wording of the ground-truth answer, and wording of distractor options. These perturbations are designed to preserve the original semantic meaning of the questions and choices, offering alternative representations of the same information. We implement these modifications both individually and in combination. This allows us to assess their impact on the performance of LALMs. Figure \ref{fig:metodology} visually represents the evaluation strategy adopted when these perturbations are applied individually.

\begin{table}[!ht]
    \caption{Number of versions evaluated per perturbation.}
    \centering
    \resizebox{0.9\linewidth}{!}{
    \begin{tabular}{l c}
        \toprule
        \textbf{Perturbation} & \textbf{Number of versions} \\
        \midrule
        Choice ordering & 24 \\
        Question rephrasing & 7 \\
        Ground truth answer rephrasing & 7 \\
        Distractor rephrasing & 7 \\
        Mix of perturbations & 7 \\
        \bottomrule
    \end{tabular}
    }
    \label{tab:number_of_perturbations}
\end{table}

Table \ref{tab:number_of_perturbations} details the number of versions used for each specific perturbation. Each perturbation type includes the default data. For example, among the 24 possible orderings of choices, one configuration reflects the default arrangement.

\subsection{Ordering of the choices}
In the MCQA scenario, LALM is tasked with selecting the correct option from a set of choices. Usually, this involves choosing from four options: one correct answer (ground truth) and three distractors. In our study, we assess the impact of different choice arrangements by evaluating all 24 possible permutations of these options.

\subsection{Rephrasing}
We rephrased the question, the correct answer, and the distractors, using two models: gemini-2.5-flash \cite{comanici2025gemini} and gemma-3-12b-it \cite{gemmateam2025gemma3technicalreport}. The gemini-2.5-flash model was prompted to generate three distinct versions for each component, using comprehensive contextual information. For example, while rephrasing distractors, the model received both the question and the correct answer as part of the prompt. On the other hand, the gemma-3-12b-it model was used to produce three paraphrased versions of the question, ground-truth answer, and distractors with varying levels of contextual information, either including or excluding the question, correct answer, and distractors.

We end up with six variations for each element in addition to the original phrasing, resulting in seven distinct versions of the benchmark for evaluation. Understanding the critical importance of paraphrasing, especially for the ground-truth answer, we manually validated 620 randomly selected samples across all benchmarks and perturbations. All preserved the original meaning, confirming the accuracy of the paraphrased answers.

\begin{table*}[h!]
\caption{Accuracy (\%) with the default setup across three benchmarks: MMAU (test-mini), MMAR, and MMSU. We include random choice and text-only LLM controls (question+options only, no audio) to estimate performance attributable to textual priors.}
\vspace{-0.2cm}
\centering
\resizebox{\textwidth}{!}{
\begin{tabular}{l|cccc|cccc|ccc}
\hline
\multirow{2}{*}{\textbf{Model}} 
 & \multicolumn{4}{c|}{\textbf{MMAU (test-mini)}} 
 & \multicolumn{4}{c|}{\textbf{MMAR}} 
 & \multicolumn{3}{c}{\textbf{MMSU}} \\
\cline{2-12}
 & \textbf{Sound} & \textbf{Music} & \textbf{Speech} & \textbf{Avg}
 & \textbf{Sound} & \textbf{Music} & \textbf{Speech} & \textbf{Avg}
 & \textbf{Perception} & \textbf{Reasoning} & \textbf{Avg} \\
\hline
Random choice          & 28.5 & 23.7 & 24.9 & 25.7  & 34.6 & 26.2 & 36.7 & 33.0  & 26.4 & 25.0 & 25.4 \\
\hline
Qwen2.5-7B-Instruct    & 37.5 & 34.4 & 32.4 & 34.8 & \textbf{32.7} & 29.1 & 36.4 & 32.2 & 55.4 & 31.4 & 37.8 \\
Llama-3.1-8B-Instruct  & 41.1 & 41.0 & 30.3 & 37.5 & 26.1 & 35.0 & 34.4 & 32.4 & \textbf{56.4} & 31.7 & 38.4 \\
gemma-3-27B-it         & \textbf{50.2} & \textbf{47.1} & \textbf{47.8} & \textbf{48.3} & 30.3 & \textbf{40.8} & \textbf{41.2} & \textbf{35.8} & 29.2 & \textbf{49.3} & \textbf{38.9} \\
\hline
LTU-AS \cite{sakshi2024mmau, ma2025mmar, wang2025mmsu} & 23.4 & 9.1 & 20.6 & 18.9 & 20.0 & 14.1 & 19.1 & 19.0 & 24.1 & 25.9 & 25.0\\
SALMONN 13B \cite{sakshi2024mmau, ma2025mmar} & 41.0 & 34.8 & 25.5 & 33.7 & 30.3 & 31.1 & 34.7 & 33.2 & - & - & - \\
Qwen-Audio-Chat \cite{sakshi2024mmau, ma2025mmar, wang2025mmsu} & 55.3 & 44.0 & 30.0 & 43.1 & 27.9 & 20.4 & 22.1 & 23.5 & 35.7 & 55.9 & 46.9 \\
Audio Flamingo 2       & 70.6 & \textbf{72.5} & 45.1 & 62.7  & 47.3 & 43.7 & 45.9 & 45.3  & 32.6 & 51.5 & 41.8 \\
Audio Flamingo 3       & \textbf{80.2} & 74.0 & 65.8 & \textbf{73.3}  
                       & 53.3 & 49.5 & 62.6 & 58.5  
                       & 46.4 & 76.6 & 61.0 \\
Qwen2.5-Omni-7B        & 71.8 & 67.1 & 65.8 & 68.2  
                       & \textbf{60.6} & 42.2 & \textbf{62.6} & \textbf{59.0}  
                       & 47.6 & \textbf{78.1} & \textbf{62.4} \\
Kimi-Audio-7B-Instruct & 76.3 & 67.1 & \textbf{70.0} & 71.1  
                       & 52.1 & 43.2 & 58.5 & 54.0  
                       & 46.1 & 71.5 & 58.4 \\
\hline
\end{tabular}
}
\label{tab:default_eval}
\end{table*}

\subsection{Mix of perturbations}
When combining all possible perturbations, we assign a probability of 0.5 to the application of each modification to an individual triplet of question, ground-truth answer, and distractors. This approach leads to a low likelihood of either applying all modifications or none, precisely quantified at 0.0625.

\subsection{Evaluation metrics}
We evaluate performance using the following measures. The first is the accuracy through mean, standard deviation, minimum, and maximum values. Furthermore, we calculate the consistency rate (CR), as described in \cite{nalbandyan-etal-2025-score}, and the CoR to gauge the reliability and correctness of the results. CoR and CR capture complementary aspects of the model behavior.

Let $Q = \{Q_1, Q_2, \dots, Q_n\}$ denote the set of questions, where $n$ represents the total number of triplet benchmarks (question, audio, and choices). Correspondingly, $R = \{R_1, R_2, \ldots, R_n\}$ represents the collection of response sets, with each $R_i = \{r_{i1}, r_{i2}, \ldots, r_{im}\}$ indicating responses for individual questions, where $m$ is the number of different perturbations evaluated. $G = \{g_1, g_2, \ldots, g_n\}$ represents the list of ground-truth answers.

\noindent \textbf{Consistency rate (CR)} is defined as:
\begin{equation}
    \text{CR} = \frac{1}{n} \sum_{i=1}^{n} 
        \sum_{j=1}^{m} \sum_{k=j+1}^{m} \frac{\delta(r_{ij}, r_{ik})}{\binom{m}{2}}.
\end{equation}
where $\delta(x, y)$ is the Kronecker delta,
\begin{equation}
\delta(x, y) =
\begin{cases}
1 & \text{if } x = y, \\
0 & \text{otherwise}.
\end{cases}
\end{equation}
CR assesses internal robustness by examining the agreement among multiple responses. CR is applicable when ground-truth answers or distractors remain unchanged.

\noindent \textbf{Correctness rate (CoR)} is defined as:
\begin{equation}
    \text{CoR} = \frac{1}{n} \sum_{i=1}^{n} \prod_{j=1}^{m} \text{correctness}(r_{ij}, g_i),
\end{equation}
where $\text{correctness}(r, g)$ is given by the benchmark. It returns a value of 1 if the answer is correct, and 0 otherwise. Some benchmarks use a correctness function based on a regular expression, whereas others require an exact match of content. CoR counts a question as correct only if it is answered correctly under all perturbations, making it stricter than mean accuracy. This metric applies to all perturbations in our methodology.

\section{Experiments}
\label{sec:experiments}
In our experimental setup, we have developed perturbed versions of the benchmark that preserve the original audio while modifying only the text input provided to the LALM. For the choice-ordering permutation, the original text length is preserved. However, when rephrasing the questions, we noticed a decrease in average length, with the paraphrased questions containing approximately 9.6 words compared to the original average of 12.5 words. In contrast, the length of paraphrased choices tends to increase, averaging 6.1 words per choice compared to 4.3 words in the original text.

\subsection{Benchmarks}
We employ three benchmarks to evaluate LALMs across diverse audio domains. The MMAU-v05.15.25 (test-mini) subset \cite{sakshi2024mmau} contains 1k samples spanning speech, sound, and music, providing a broad measure of audio understanding. MMAR \cite{ma2025mmar}, with 1k samples across speech, sound, music, and mixed-source categories, introduces more complex multi-source scenarios. MMSU \cite{wang2025mmsu}, focused on speech and comprising 5k samples, enables evaluation of reasoning in linguistically rich contexts. Together, these datasets offer a comprehensive basis for assessing both general and specialized reasoning in LALMs.

\subsection{Large Audio Language Models}
We conducted evaluations on four open-source LALMs. Audio Flamingo 2 (3.2B) \cite{ghoshaudio} based on cross-attention mechanism, while the remaining three models are based on self-attention mechanisms: Audio Flamingo 3 (8.4B) \cite{goel2025audioflamingo3advancing}, Qwen2.5-Omni-7B \cite{xu2025qwen2}, and Kimi-Audio-7B-Instruct \cite{ding2025kimi}.

We perform a greedy decoding strategy and use a standardized prompt across all models, directing the model to return the selected candidate in the format: \texttt{(<Choice ID>) <Text>}.

\subsection{Input and output processing}
Generally, benchmarks consist of four options per question. However, certain questions may present only two options (e.g. yes/no) or exceed four options. In our experiments, we reduced the choices to four while ensuring the inclusion of the ground-truth answer among these selections.

\begin{table*}[h!]
\caption{Effect of perturbations across benchmarks (MMAU, MMAR, MMSU). ACC\% shows the accuracy mean$\pm$std, with a range of min--max. CR is the consistency rate, and CoR denotes the correctness rate.}
\vspace{-0.2cm}
\resizebox{\textwidth}{!}{
\begin{tabular}{l c c c c c c c c c}
\toprule
\multirow{2}{*}{\textbf{Model}} 
  & \multicolumn{3}{c}{\textbf{MMAU}} 
  & \multicolumn{3}{c}{\textbf{MMAR}} 
  & \multicolumn{3}{c}{\textbf{MMSU}} \\
\cmidrule(lr){2-4} \cmidrule(lr){5-7} \cmidrule(lr){8-10}
 & \textbf{ACC\% (mean$\pm$std, [min, max])} & \textbf{CR} & \textbf{CoR} 
 & \textbf{ACC\% (mean$\pm$std, [min, max])} & \textbf{CR} & \textbf{CoR} 
 & \textbf{ACC\% (mean$\pm$std, [min, max])} & \textbf{CR} & \textbf{CoR} \\
\midrule

\multicolumn{10}{c}{\textbf{Choice ordering permutation}} \\
\midrule
Audio Flamingo 2        & 58.9 $\pm$ 1.4 {[57.3, 63.0]} & 0.80 & 0.43
                        & 44.9 $\pm$ 0.8 {[43.0, 45.9]} & 0.74 & 0.21
                        & 41.2 $\pm$ 1.2 {[38.2, 42.8]} & 0.63 & 0.15 \\
Audio Flamingo 3        & 72.8 $\pm$ 0.6 {[71.4, 74.0]} & 0.88 & \textbf{0.63}
                        & 58.0 $\pm$ 0.8 {[56.9, 60.1]} & 0.84 & \textbf{0.40}
                        & 60.8 $\pm$ 0.4 {[60.2, 61.4]} & 0.76 & \textbf{0.42} \\
Qwen2.5-Omni-7B         & 68.1 $\pm$ 1.4 {[65.3, 70.6]} & 0.76 & 0.49
                        & 57.4 $\pm$ 0.9 {[55.7, 58.8]} & 0.74 & 0.36 
                        & 62.5 $\pm$ 0.3 {[62.0, 63.4]} & 0.63 & 0.40 \\
Kimi-Audio-7B-Instruct  & 63.8 $\pm$ 2.7 {[58.9, 71.1]} & 0.69 & 0.33
                        & 54.5 $\pm$ 0.9 {[53.2, 56.7]} & 0.65 & 0.25
                        & 58.5 $\pm$ 0.6 {[57.5, 59.8]} & 0.64 & 0.27 \\
\midrule

\multicolumn{10}{c}{\textbf{Question rephrasing}} \\
\midrule
Audio Flamingo 2        & 62.3 $\pm$ 0.5 [61.6, 63.0] & 0.86 & 0.52
                        & 45.6 $\pm$ 0.8 [44.7, 46.9] & 0.80 & 0.29
                        & 41.3 $\pm$ 1.4 [39.1, 43.7] & 0.79 & 0.27 \\
Audio Flamingo 3        & 73.1 $\pm$ 0.7 [71.7, 74.1] & 0.90 & \textbf{0.65}
                        & 57.5 $\pm$ 0.8 [55.9, 58.5] & 0.83 & 0.43
                        & 60.7 $\pm$ 1.1 [59.5, 62.8] & 0.84 & 0.48 \\
Qwen2.5-Omni-7B         & 66.5 $\pm$ 0.8 [65.5, 67.6] & 0.85 & 0.59
                        & 56.2 $\pm$ 0.5 [55.4, 56.9] & 0.83 & \textbf{0.44}
                        & 62.2 $\pm$ 0.8 [60.8, 63.7] & 0.79 & \textbf{0.51} \\
Kimi-Audio-7B-Instruct  & 68.9 $\pm$ 1.1 [67.6, 71.1] & 0.86 & 0.58
                        & 51.4 $\pm$ 1.1 [50.2, 54.0] & 0.81 & 0.36
                        & 57.1 $\pm$ 1.2 [55.9, 59.0] & 0.81 & 0.42 \\
\midrule

\multicolumn{10}{c}{\textbf{Ground-truth answer rephrasing}} \\
\midrule
Audio Flamingo 2        & 70.1 $\pm$ 5.1 [63.0, 78.2] & - & 0.42 
                        & 63.2 $\pm$ 10.2 [45.5, 72.6] & - & 0.30 
                        & 61.2 $\pm$ 9.8 [41.9, 70.3] & - & 0.28 \\
Audio Flamingo 3        & 77.3 $\pm$ 4.1 [72.7, 85.5] & - & 0.54 
                        & 70.5 $\pm$ 7.1 [58.5, 77.0] & - & \textbf{0.41} 
                        & 69.1 $\pm$ 6.6 [57.7, 75.7] & - & 0.43 \\
Qwen2.5-Omni-7B         & 64.9 $\pm$ 2.9 [59.1, 68.5] & - & 0.33 
                        & 61.6 $\pm$ 5.1 [51.0, 65.8] & - & 0.33 
                        & 68.8 $\pm$ 5.8 [58.3, 76.0] & - & \textbf{0.44} \\
Kimi-Audio-7B-Instruct  & 77.1 $\pm$ 5.4 [69.1, 86.0] & - & \textbf{0.57} 
                        & 65.8 $\pm$ 8.2 [53.3, 74.6] & - & 0.40 
                        & 66.4 $\pm$ 6.1 [55.5, 72.0] & - & 0.38 \\
\midrule

\multicolumn{10}{c}{\textbf{Distractors rephrasing}} \\
\midrule
Audio Flamingo 2        & 42.7 $\pm$ 13.7 [27.1, 63.0] & - & 0.21 
                        & 27.9 $\pm$ 8.8 [16.0, 45.5] & - & 0.07 
                        & 28.5 $\pm$ 6.3 [20.0, 35.1] & - & 0.10 \\
Audio Flamingo 3        & 58.8 $\pm$ 10.4 [46.4, 73.3] & - & 0.42 
                        & 42.4 $\pm$ 8.8 [30.4, 58.5] & - & 0.21 
                        & 53.6 $\pm$ 5.9 [46.5, 61.1] & - & 0.36 \\
Qwen2.5-Omni-7B         & 58.9 $\pm$ 5.3 [52.0, 66.0] & - & \textbf{0.50} 
                        & 51.5 $\pm$ 3.9 [47.0, 57.0] & - & \textbf{0.34} 
                        & 60.7 $\pm$ 3.5 [57.2, 66.1] & - & \textbf{0.46} \\
Kimi-Audio-7B-Instruct  & 54.5 $\pm$ 10.8 [42.1, 71.1] & - & 0.37 
                        & 42.4 $\pm$ 7.0 [33.5, 54.0] & - & 0.23 
                        & 52.8 $\pm$ 4.7 [47.3, 59.4] & - & 0.32 \\

\midrule
\multicolumn{10}{c}{\textbf{Mix of perturbations}} \\
\midrule
Audio Flamingo 2        & 56.8 $\pm$ 3.0 {[53.7, 63.0]} & - & 0.23
                        & 46.1 $\pm$ 0.8 {[44.9, 47.0]} & - & 0.09
                        & 45.0 $\pm$ 1.4 {[41.9, 46.5]} & - & 0.12 \\
Audio Flamingo 3        & 69.7 $\pm$ 1.7 {[67.5, 73.3]} & - & \textbf{0.45}
                        & 55.9 $\pm$ 1.2 {[54.5, 58.5]} & - & 0.21
                        & 61.8 $\pm$ 0.3 {[61.4, 62.5]} & - & 0.32 \\
Qwen2.5-Omni-7B         & 63.4 $\pm$ 1.4 {[62.2, 66.0]} & - & 0.35
                        & 56.1 $\pm$ 1.0 {[54.8, 57.7]} & - & \textbf{0.26}
                        & 64.1 $\pm$ 0.8 {[62.5, 65.1]} & - & \textbf{0.37} \\
Kimi-Audio-7B-Instruct  & 64.6 $\pm$ 2.9 {[61.5, 71.1]} & - & 0.32
                        & 53.7 $\pm$ 0.8 {[52.2, 54.8]} & - & 0.21
                        & 59.4 $\pm$ 0.5 {[58.7, 60.2]} & - & 0.25 \\

\bottomrule
\end{tabular}
}
\label{tab:results_permutations}
\end{table*}

The Kimi-Audio-7B-Instruct model exclusively outputs the chosen letter for the selected option. To use this model for evaluation purposes, we parse its output to conform to the standardized format (the chosen letter and answer content).

\subsection{Text-only controls}
To contextualize LALM accuracy, we report results for three text-only LLMs (Qwen2.5-7B-Instruct \cite{qwen2025qwen25technicalreport}, Llama-3.1-8B-Instruct \cite{grattafiori2024llama}, and gemma-3-27B-it \cite{gemmateam2025gemma3technicalreport}). Models receive the same questions and answer options, but no audio. This estimates how much of each benchmark can be solved via textual priors. This language bias can come from lexical cues in the question/options, stereotypical answer frequency, or artifacts introduced by option formatting rather than audio understanding. For evaluation, we use the same formatting and parsing protocol as the LALMs, enabling direct cross-modal comparison.

\section{Results and Analysis}
\label{sec:typestyle}
In Table \ref{tab:default_eval}, we present the default setup performance metrics for the four LALMs under evaluation. This setup respects the original order of answer choices and phrasing of questions, ground truth answers, and distractors. The results closely align with those reported in previous studies \cite{sakshi2024mmau, ma2025mmar, wang2025mmsu}, showing a small accuracy drift of approximately $\pm 2$\%, potentially attributable to variances in decoding strategies or library versions. For broader context, we additionally include published results for three reference models: LTU-AS 7B \cite{gong2023joint}, SALMONN 13B \cite{tang2024salmonn}, and Qwen-Audio-Chat 8.4B \cite{chu2023qwen}.

Table \ref{tab:default_eval} further reports performance for a random choice baseline and for text-only LLMs evaluated without any audio input. Notably, in the MMAR benchmark, the random choice accuracy is justified by the 17.1\% of the questions with only two options. Additionally, the text-only LLM results reveal that existing MCQA benchmarks may contain textual shortcuts exploitable without any audio signal. On both MMAU and MMSU, all evaluated text-only LLMs exceed random chance accuracy. More notably, gemma-3-27B-it achieves 48.3\% accuracy on MMAU (22.6 percentage points above the random baseline) and surpasses both LTU-AS and SALMONN across all benchmarks. For Qwen-Audio-Chat, the text-only models outperform the audio model on MMAU and MMAR.  Considering all this, random choice should not be treated as a baseline for LALMs, and text-only is a meaningful baseline to isolate genuine audio understanding from potential language bias.

Table \ref{tab:results_permutations} summarizes the model performance in the three benchmarks studied under different types of perturbations. Both choice ordering and question rephrasing resulted in moderate variability, suggesting that the models have some capacity to deal with these perturbations. In contrast, rephrasing the ground-truth answer caused a higher variance, and distractor rephrasing resulted in the biggest drop in both accuracy and CoR. This shows that models are highly sensitive to the wording of choices. When perturbations are combined, we can effectively capture the variability introduced by each perturbation moderately. This is well represented in the CoR. 

Model-wise, Audio Flamingo 3 achieves the highest overall accuracy. It remains robust under choice ordering and question rephrasing, also showing relatively strong CR and CoR. Qwen2.5-Omni-7B performs slightly below Audio Flamingo 3 in accuracy, but demonstrates stronger stability against distractor rephrasing, where it achieves the highest CoR across benchmarks. Kimi-Audio-7B-Instruct performs competitively under ground-truth answer rephrasing, occasionally matching or surpassing Audio Flamingo 3. Nevertheless, it is largely affected by distractor rephrasing and the mix of perturbations. Audio Flamingo 2, the smallest model without integrated speech in its training, consistently ranks lowest across all perturbative conditions, showing particularly weak resilience when confronted with distractor rephrasing. Notably, CoR captures what mean accuracy misses, models with similar averages can differ substantially in robustness to perturbations. 


Additionally, paraphrasing choices increases variability in accuracy. Given that the paraphrased versions of choices typically result in longer text than the original (see Section~\ref{sec:experiments}), we hypothesize that the cause is a bias toward longer options. To test this, we measured the proportion of times each model selected the longest choice. Audio Flamingo 2 chose it 52.38\% of the time, Audio Flamingo 3 50.31\%, Kimi-Audio 50.23\%, and Qwen2.5-Omni-7B 42.51\%. While in data, the longest choice is the correct one for the 45.05\% of the samples. This tendency was amplified when the longest choice was correct, rising to 70.93\%, 77.94\%, 76.41\%, and 74.29\%, respectively, suggesting 
that LALMs rely on choice length as a selection shortcut. 

Exhaustively applying all possible perturbation combinations would be computationally prohibitive. We therefore recommend the mix of perturbations as an evaluation protocol, scored via CoR. It captures sensitivity to both rephrasing and choice ordering while remaining practical for robust LALM evaluation within the MCQA paradigm.

\section{Limitations and Future Work}
We isolate linguistic robustness by keeping the audio signal constant; signal-level perturbations remain a complementary future direction. The proposed perturbations could also serve as data augmentation to mitigate linguistic bias during training.

\section{Conclusions}
\label{sec:conclusions}
MCQA evaluation for LALMs, while useful, presents some limitations. Our findings show that accuracy can be inflated by language bias, making text-only controls meaningful as a baseline to isolate genuine audio understanding. Moreover, LALMs are highly sensitive to minor perturbations in MCQA, with distractor rephrasings causing the largest performance drops. Audio Flamingo 3 achieves the best overall accuracy, while Qwen2.5-Omni-7B shows stronger robustness to distractor wording. We also find a bias of LALMs toward selecting longer choices. To address this, we propose a robustness evaluation framework that mixes perturbations. Thus, a limited number of perturbations effectively reflects models' limitations. This, together with the correctness rate, offers a reliable measure of robustness.

\section{Generative AI Use Disclosure}
We used a generative AI tool to paraphrase and polish portions of the manuscript to improve readability and grammar.


\section{Acknowledgments}
\label{sec:ack}
This project has been partially funded by the European Union’s Horizon 2020 RIA ELOQUENCE project (Grant Agreement No. 101135916). Views and opinions expressed are, however, those of the author(s) only and do not necessarily reflect those of the European Union or European Commission-EU. Neither the European Union nor the granting authority can be held responsible for them.
Santosh Kesiraju is supported by the Ministry of Education, Youth and Sports of the Czech Republic (MoE) through the OP JAK project ``Linguistics, Artificial Intelligence and Language and Speech Technologies: from Research to Applications'' (ID:CZ.02.01.01/00/23\_020/0008518). Part of the work was conducted during the JSALT 2025 Workshop.


\begin{thebibliography}{10}
\providecommand{\url}[1]{#1}
\csname url@samestyle\endcsname
\providecommand{\newblock}{\relax}
\providecommand{\bibinfo}[2]{#2}
\providecommand{\BIBentrySTDinterwordspacing}{\spaceskip=0pt\relax}
\providecommand{\BIBentryALTinterwordstretchfactor}{4}
\providecommand{\BIBentryALTinterwordspacing}{\spaceskip=\fontdimen2\font plus
\BIBentryALTinterwordstretchfactor\fontdimen3\font minus \fontdimen4\font\relax}
\providecommand{\BIBforeignlanguage}[2]{{%
\expandafter\ifx\csname l@#1\endcsname\relax
\typeout{** WARNING: IEEEtran.bst: No hyphenation pattern has been}%
\typeout{** loaded for the language `#1'. Using the pattern for}%
\typeout{** the default language instead.}%
\else
\language=\csname l@#1\endcsname
\fi
#2}}
\providecommand{\BIBdecl}{\relax}
\BIBdecl

\bibitem{gong2023listen}
Y.~Gong, H.~Luo, A.~H. Liu, L.~Karlinsky, and J.~Glass, ``Listen, think, and understand,'' in \emph{Proc. of ICLR}, 2024.

\bibitem{sun2024video}
G.~Sun, W.~Yu, C.~Tang, X.~Chen, T.~Tan, W.~Li, L.~Lu, Z.~Ma, Y.~Wang, and C.~Zhang, ``video-salmonn: Speech-enhanced audio-visual large language models,'' in \emph{Proc. of ICML}, 2024.

\bibitem{ghosh2024gama}
\BIBentryALTinterwordspacing
S.~Ghosh, S.~Kumar, A.~Seth, C.~K.~R. Evuru, U.~Tyagi, S.~Sakshi, O.~Nieto, R.~Duraiswami, and D.~Manocha, ``{GAMA: A large audio-language model with advanced audio understanding and complex reasoning abilities},'' in \emph{Proc. of EMNLP}.\hskip 1em plus 0.5em minus 0.4em\relax Miami, Florida, USA: ACL, 2024, pp. 6288--6313. [Online]. Available: \url{https://aclanthology.org/2024.emnlp-main.361/}
\BIBentrySTDinterwordspacing

\bibitem{ghoshaudio}
S.~Ghosh, Z.~Kong, S.~Kumar, S.~Sakshi, J.~Kim, W.~Ping, R.~Valle, D.~Manocha, and B.~Catanzaro, ``{Audio Flamingo 2: An Audio-Language Model with Long-Audio Understanding and Expert Reasoning Abilities},'' in \emph{Proc. of ICML}, 2025.

\bibitem{xu2025qwen2}
J.~Xu, Z.~Guo, J.~He, H.~Hu, T.~He, S.~Bai, K.~Chen, J.~Wang, Y.~Fan, K.~Dang \emph{et~al.}, ``Qwen2. 5-omni technical report,'' \emph{arXiv preprint arXiv:2503.20215}, 2025.

\bibitem{xie2025audio}
X.~Zhifei, M.~Lin, Z.~Liu, P.~Wu, S.~Yan, and C.~Miao, ``Audio-reasoner: Improving reasoning capability in large audio language models,'' in \emph{Proceedings of the 2025 Conference on Empirical Methods in Natural Language Processing}, 2025, pp. 23\,840--23\,862.

\bibitem{ding2025kimi}
D.~Ding, Z.~Ju, Y.~Leng, S.~Liu, T.~Liu, Z.~Shang, K.~Shen, W.~Song, X.~Tan, H.~Tang \emph{et~al.}, ``Kimi-audio technical report,'' \emph{arXiv preprint arXiv:2504.18425}, 2025.

\bibitem{goel2025audioflamingo3advancing}
\BIBentryALTinterwordspacing
S.~Ghosh, A.~Goel, J.~Kim, S.~Kumar, Z.~Kong, S.~gil Lee, C.-H.~H. Yang, R.~Duraiswami, D.~Manocha, R.~Valle, and B.~Catanzaro, ``Audio flamingo 3: Advancing audio intelligence with fully open large audio language models,'' in \emph{The Thirty-ninth Annual Conference on Neural Information Processing Systems}, 2025. [Online]. Available: \url{https://openreview.net/forum?id=FjByDpDVIO}
\BIBentrySTDinterwordspacing

\bibitem{huang2024dynamic}
C.-y. Huang, K.-H. Lu, S.-H. Wang, C.-Y. Hsiao, C.-Y. Kuan, H.~Wu, S.~Arora, K.-W. Chang, J.~Shi, Y.~Peng \emph{et~al.}, ``Dynamic-superb: Towards a dynamic, collaborative, and comprehensive instruction-tuning benchmark for speech,'' in \emph{Proc. of ICASSP}.\hskip 1em plus 0.5em minus 0.4em\relax IEEE, 2024, pp. 12\,136--12\,140.

\bibitem{yang2024air}
\BIBentryALTinterwordspacing
Q.~Yang, J.~Xu, W.~Liu, Y.~Chu, Z.~Jiang, X.~Zhou, Y.~Leng, Y.~Lv, Z.~Zhao, C.~Zhou \emph{et~al.}, ``Air-bench: Benchmarking large audio-language models via generative comprehension,'' in \emph{Proc. of ACL}.\hskip 1em plus 0.5em minus 0.4em\relax Bangkok, Thailand: ACL, Aug. 2024, pp. 1979--1998. [Online]. Available: \url{https://aclanthology.org/2024.acl-long.109/}
\BIBentrySTDinterwordspacing

\bibitem{wang2024audiobench}
B.~Wang, X.~Zou, G.~Lin, S.~Sun, Z.~Liu, W.~Zhang, Z.~Liu, A.~Aw, and N.~F. Chen, ``Audiobench: A universal benchmark for audio large language models,'' in \emph{Proc. of NAACL:HLT}, Albuquerque, New Mexico, 2025, pp. 4297--4316.

\bibitem{huang2024dynamic2}
C.-y. Huang, W.-C. Chen, S.-w. Yang, A.~T. Liu, C.-A. Li, Y.-X. Lin, W.-C. Tseng, A.~Diwan, Y.-J. Shih, J.~Shi \emph{et~al.}, ``Dynamic-superb phase-2: A collaboratively expanding benchmark for measuring the capabilities of spoken language models with 180 tasks,'' in \emph{Proc. of ICLR}, 2025.

\bibitem{sakshi2024mmau}
S.~Sakshi, U.~Tyagi, S.~Kumar, A.~Seth, R.~Selvakumar, O.~Nieto, R.~Duraiswami, S.~Ghosh, and D.~Manocha, ``{MMAU: A massive multi-task audio understanding and reasoning benchmark},'' in \emph{Proc. of ICLR}, 2024.

\bibitem{ma2025mmar}
Z.~Ma, Y.~Ma, Y.~Zhu, C.~Yang, Y.-W. Chao, R.~Xu, W.~Chen, Y.~Chen, Z.~Chen, J.~Cong \emph{et~al.}, ``Mmar: A challenging benchmark for deep reasoning in speech, audio, music, and their mix,'' in \emph{The Thirty-ninth Annual Conference on Neural Information Processing Systems Datasets and Benchmarks Track}, 2025.

\bibitem{yang2025sakura}
C.-K. Yang, N.~Ho, Y.-T. Piao, and H.~yi~Lee, ``{SAKURA: On the Multi-hop Reasoning of Large Audio-Language Models Based on Speech and Audio Information},'' in \emph{{Interspeech 2025}}, 2025, pp. 1788--1792.

\bibitem{wang2025mmsu}
D.~Wang, J.~Wu, J.~Li, D.~Yang, X.~Chen, T.~Zhang, and H.~Meng, ``{MMSU: A Massive Multi-task Spoken Language Understanding and Reasoning Benchmark},'' \emph{arXiv preprint arXiv:2506.04779}, 2025.

\bibitem{kumar2026mmau}
S.~Kumar, {\v{S}}.~Sedl{\'a}{\v{c}}ek, V.~Lokegaonkar, F.~L{\'o}pez, W.~Yu, N.~Anand, H.~Ryu, L.~Chen, M.~Pli{\v{c}}ka, M.~Hlav{\'a}{\v{c}}ek \emph{et~al.}, ``Mmau-pro: A challenging and comprehensive benchmark for holistic evaluation of audio general intelligence,'' in \emph{Proceedings of the AAAI Conference on Artificial Intelligence}, vol.~40, no.~27, 2026, pp. 22\,688--22\,697.

\bibitem{zheng:2024:LLM_MCQ}
\BIBentryALTinterwordspacing
C.~Zheng, H.~Zhou, F.~Meng, J.~Zhou, and M.~Huang, ``{Large Language Models are not Robust Multiple Choice Selectors},'' in \emph{Proc.of ICLR}, 2024. [Online]. Available: \url{https://openreview.net/forum?id=shr9PXz7T0}
\BIBentrySTDinterwordspacing

\bibitem{hendrycks2020measuring}
D.~Hendrycks, C.~Burns, S.~Basart, A.~Zou, M.~Mazeika, D.~Song, and J.~Steinhardt, ``Measuring massive multitask language understanding,'' in \emph{Proc. of ICLR}, 2021.

\bibitem{zellers2019hellaswag}
\BIBentryALTinterwordspacing
R.~Zellers, A.~Holtzman, Y.~Bisk, A.~Farhadi, and Y.~Choi, ``Hellaswag: Can a machine really finish your sentence?'' in \emph{Proc. of ACL}.\hskip 1em plus 0.5em minus 0.4em\relax Florence, Italy: ACL, Jul. 2019, pp. 4791--4800. [Online]. Available: \url{https://aclanthology.org/P19-1472/}
\BIBentrySTDinterwordspacing

\bibitem{wang2024mmlu}
\BIBentryALTinterwordspacing
Y.~Wang, X.~Ma, G.~Zhang, Y.~Ni, A.~Chandra, S.~Guo, W.~Ren, A.~Arulraj, X.~He, Z.~Jiang, T.~Li, M.~Ku, K.~Wang, A.~Zhuang, R.~Fan, X.~Yue, and W.~Chen, ``{MMLU}-pro: A more robust and challenging multi-task language understanding benchmark,'' in \emph{The Thirty-eight Conference on Neural Information Processing Systems Datasets and Benchmarks Track}, 2024. [Online]. Available: \url{https://openreview.net/forum?id=y10DM6R2r3}
\BIBentrySTDinterwordspacing

\bibitem{balepur-etal-2025-best}
\BIBentryALTinterwordspacing
N.~Balepur, R.~Rudinger, and J.~L. Boyd-Graber, ``Which of these best describes multiple choice evaluation with {LLM}s? a) forced {B}) flawed {C}) fixable {D}) all of the above,'' in \emph{Proc. of ACL (Volume 1: Long Papers)}.\hskip 1em plus 0.5em minus 0.4em\relax Vienna, Austria: ACL, Jul. 2025, pp. 3394--3418. [Online]. Available: \url{https://aclanthology.org/2025.acl-long.169/}
\BIBentrySTDinterwordspacing

\bibitem{nalbandyan-etal-2025-score}
\BIBentryALTinterwordspacing
G.~Nalbandyan, R.~Shahbazyan, and E.~Bakhturina, ``{SCORE}: Systematic {CO}nsistency and robustness evaluation for large language models,'' in \emph{Proc. of NAACL: HLT (Volume 3: Industry Track)}.\hskip 1em plus 0.5em minus 0.4em\relax Albuquerque, New Mexico: ACL, Apr. 2025, pp. 470--484. [Online]. Available: \url{https://aclanthology.org/2025.naacl-industry.39/}
\BIBentrySTDinterwordspacing

\bibitem{bhattacharya2025benchmarking}
D.~Bhattacharya, A.~Kulkarni, and S.~Ganapathy, ``Benchmarking and confidence evaluation of lalms for temporal reasoning,'' in \emph{Proc. Interspeech 2025}, 2025, pp. 2068--2072.

\bibitem{comanici2025gemini}
G.~Comanici, E.~Bieber, M.~Schaekermann, I.~Pasupat, N.~Sachdeva, I.~Dhillon, M.~Blistein, O.~Ram, D.~Zhang, E.~Rosen \emph{et~al.}, ``Gemini 2.5: Pushing the frontier with advanced reasoning, multimodality, long context, and next generation agentic capabilities,'' \emph{arXiv preprint arXiv:2507.06261}, 2025.

\bibitem{gemmateam2025gemma3technicalreport}
\BIBentryALTinterwordspacing
G.~Team, A.~Kamath, J.~Ferret, S.~Pathak, N.~Vieillard, R.~Merhej, S.~Perrin, T.~Matejovicova, A.~Ramé, M.~Rivière \emph{et~al.}, ``Gemma 3 technical report,'' 2025. [Online]. Available: \url{https://arxiv.org/abs/2503.19786}
\BIBentrySTDinterwordspacing

\bibitem{qwen2025qwen25technicalreport}
\BIBentryALTinterwordspacing
Qwen, :, A.~Yang, B.~Yang, B.~Zhang, B.~Hui, B.~Zheng, B.~Yu, C.~Li, D.~Liu \emph{et~al.}, ``Qwen2.5 technical report,'' 2025. [Online]. Available: \url{https://arxiv.org/abs/2412.15115}
\BIBentrySTDinterwordspacing

\bibitem{grattafiori2024llama}
A.~Grattafiori, A.~Dubey, A.~Jauhri, A.~Pandey, A.~Kadian, A.~Al-Dahle, A.~Letman, A.~Mathur, A.~Schelten, A.~Vaughan \emph{et~al.}, ``The llama 3 herd of models,'' \emph{arXiv preprint arXiv:2407.21783}, 2024.

\bibitem{gong2023joint}
Y.~Gong, A.~H. Liu, H.~Luo, L.~Karlinsky, and J.~Glass, ``Joint audio and speech understanding,'' in \emph{2023 IEEE Automatic Speech Recognition and Understanding Workshop (ASRU)}.\hskip 1em plus 0.5em minus 0.4em\relax IEEE, 2023, pp. 1--8.

\bibitem{tang2024salmonn}
\BIBentryALTinterwordspacing
C.~Tang, W.~Yu, G.~Sun, X.~Chen, T.~Tan, W.~Li, L.~Lu, Z.~MA, and C.~Zhang, ``Salmonn: Towards generic hearing abilities for large language models,'' in \emph{The Twelfth International Conference on Learning Representations}, 2024. [Online]. Available: \url{https://openreview.net/forum?id=14rn7HpKVk}
\BIBentrySTDinterwordspacing

\bibitem{chu2023qwen}
Y.~Chu, J.~Xu, X.~Zhou, Q.~Yang, S.~Zhang, Z.~Yan, C.~Zhou, and J.~Zhou, ``Qwen-audio: Advancing universal audio understanding via unified large-scale audio-language models,'' \emph{arXiv preprint arXiv:2311.07919}, 2023.

\end{thebibliography}


\end{document}